\documentclass[10pt,twocolumn]{article}

\usepackage[margin=1in,columnsep=0.25in]{geometry}
\usepackage[utf8]{inputenc}
\usepackage[T1]{fontenc}
\usepackage{times}
\usepackage{microtype}
\usepackage{amsmath,amssymb,amsthm}
\usepackage{mathtools}
\usepackage{booktabs}
\usepackage{graphicx}
\usepackage{hyperref}
\usepackage{algorithm}
\usepackage{algorithmic}
\usepackage{xcolor}
\usepackage{enumitem}
\usepackage{caption}
\usepackage{subcaption}
\usepackage{multirow}
\usepackage{natbib}
\usepackage{titlesec}
\usepackage{fancyhdr}

\hypersetup{colorlinks=true,linkcolor=blue!70!black,citecolor=green!50!black,urlcolor=blue!60!black}

\titleformat{\section}{\normalsize\bfseries}{\thesection.}{0.5em}{}
\titleformat{\subsection}{\normalsize\bfseries}{\thesubsection.}{0.5em}{}
\titleformat{\subsubsection}{\small\itshape}{\thesubsubsection.}{0.5em}{}
\titlespacing*{\section}{0pt}{1.5ex plus 0.3ex}{0.5ex}
\titlespacing*{\subsection}{0pt}{1.0ex plus 0.3ex}{0.3ex}

\bibpunct{(}{)}{;}{a}{,}{,}

\newcommand{\R}{\mathbb{R}}
\newcommand{\softmax}{\operatorname{softmax}}
\newcommand{\cosim}{\operatorname{cos\text{-}sim}}
\newcommand{\sgprior}{\Delta_{\mathrm{SG}}}
\newcommand{\triplet}[3]{\langle #1,\; #2,\; #3 \rangle}

\title{\textbf{Inference-Time Structural Reasoning for Compositional Vision-Language Understanding}}

\author{
  \textbf{Amartya Bhattacharya}\\
  Biomedical Data Science, Geisel School of Medicine\\
  Dartmouth College\\
  {\small\texttt{amartya.bhattacharya@dartmouth.edu}}
}
\date{}

\begin{document}
\maketitle
\thispagestyle{fancy}

\begin{abstract}
\small
Vision-language models (VLMs) excel at image--text retrieval yet persistently fail at \emph{compositional reasoning}, distinguishing captions that share the same words but differ in relational structure.
We present a unified evaluation and augmentation framework benchmarking four architecturally diverse VLMs , CLIP, BLIP, LLaVA, and Qwen3-VL-8B-Thinking , on the Winoground benchmark under plain and scene-graph-augmented regimes.
We introduce a dependency-based \textbf{TextSceneGraphParser} (spaCy) extracting subject--relation--object triples, and a \textbf{Graph Asymmetry Scorer} using optimal bipartite matching to inject structural relational priors.
Caption ablation experiments (subject/object masking and swapping) reveal that Qwen3-VL-8B-Thinking achieves a group score of \textbf{62.8}, far above all encoder-based models, while a proposed multi-turn SG filtering strategy further lifts it to \textbf{66.0}, surpassing prior open-source state-of-the-art.
We analyze the capability--augmentation tradeoff and find that SG augmentation benefits already-capable models while providing negligible or negative gains for weaker baselines.
Code: \href{https://github.com/amartyacodes/Inference-Time-Structural-Reasoning-for-Compositional-Vision-Language-Understanding}{https://github.com/amartyacodes/Inference-Time-Structural-Reasoning-for-Compositional-Vision-Language-Understanding}
\end{abstract}

\section{Introduction}
\label{sec:intro}

Modern VLMs achieve remarkable performance on image captioning, visual question answering, and cross-modal retrieval. Yet a fundamental question persists: do these models truly understand the \emph{compositional structure} of language grounded in visual scenes?

Consider the minimal pair: \emph{``The dog is chasing the cat''} vs.\ \emph{``The cat is chasing the dog.''} Both share identical content words; they differ solely in the agent--patient assignment. A model relying on bag-of-words statistics assigns nearly identical scores to both captions for any image, failing this basic compositional test.

\begin{figure}[t]
  \centering
  \includegraphics[width=0.49\linewidth]{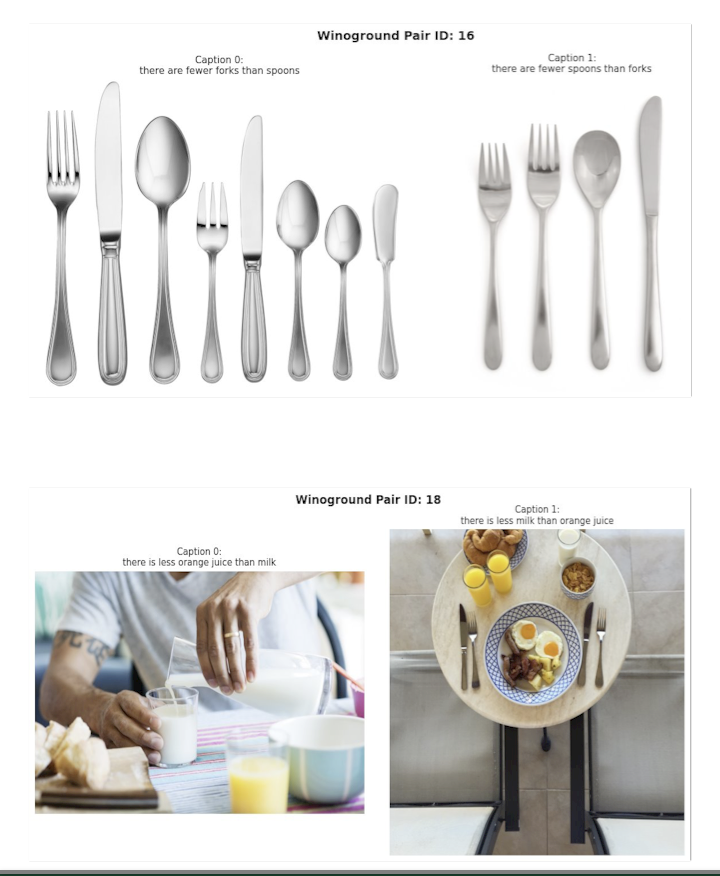}\hfill
  \includegraphics[width=0.49\linewidth]{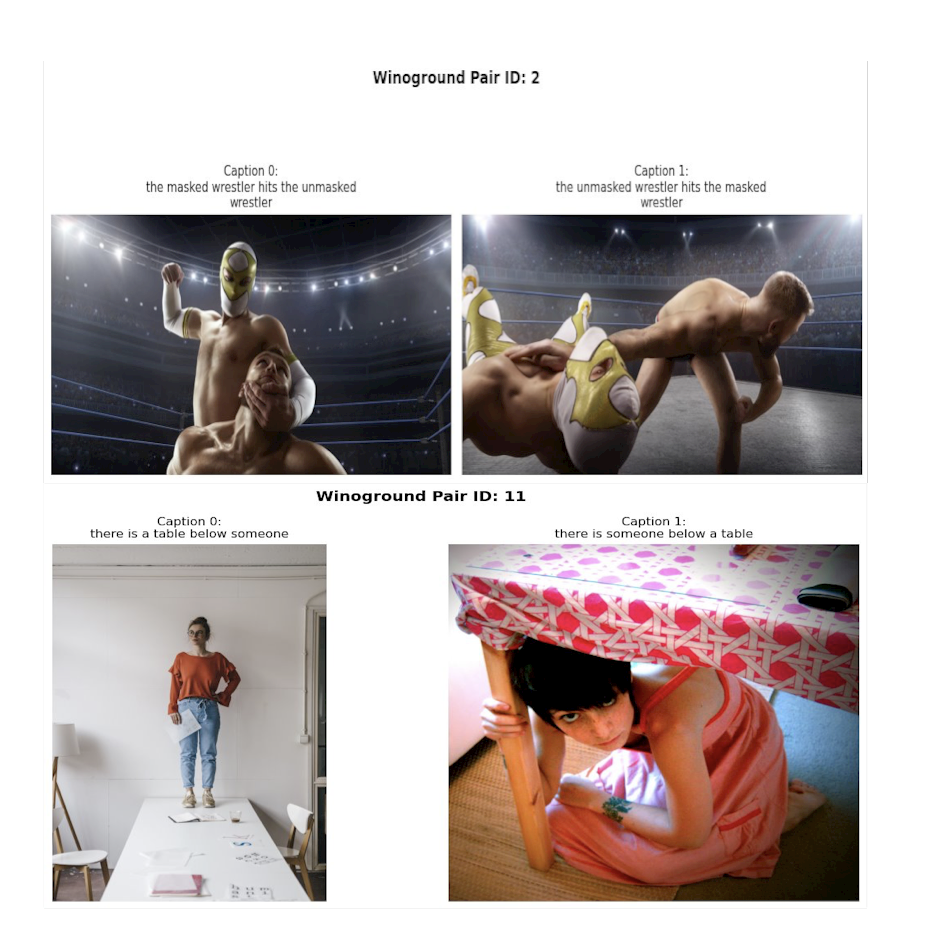}
  \caption{Winoground examples. Captions share the same words arranged differently; images share the same visual elements interacting differently. Models must correctly pair each caption with its image.}
  \label{fig:examples}
\end{figure}

The \textbf{Winoground} benchmark \citep{thrush2022winoground} operationalizes this challenge through 400 carefully curated minimally contrastive image--caption pairs (Figure~\ref{fig:examples}). Initial evaluations revealed CLIP performing only marginally above random chance, motivating the search for structural augmentation strategies.

Our work is motivated by three observations. First, \textbf{architectural diversity matters}: dual-encoder, cross-attention, and generative VLMs process compositional information through different mechanisms. Second, \textbf{explicit structure can help}: scene graphs provide a structural representation absent from raw embeddings. Third, \textbf{integration must be architecture-appropriate}: an additive score prior suits embedding models; prompt injection suits generative models.

\subsection{Contributions.}
(1)~A unified evaluation framework for four VLMs on Winoground;
(2)~a \textbf{TextSceneGraphParser} covering SVO, prepositional, copular, existential, and possessive constructions;
(3)~a \textbf{Graph Asymmetry Scorer} via optimal bipartite matching targeting agent--patient role reversal;
(4)~systematic \textbf{caption ablation experiments} quantifying entity-identity vs.\ relational-syntax reliance;
(5)~a \textbf{multi-turn SG filtering} strategy achieving group score 66.0, a new open-source state-of-the-art on Winoground.

\begin{figure*}[t]
  \centering
  \includegraphics[width=\linewidth]{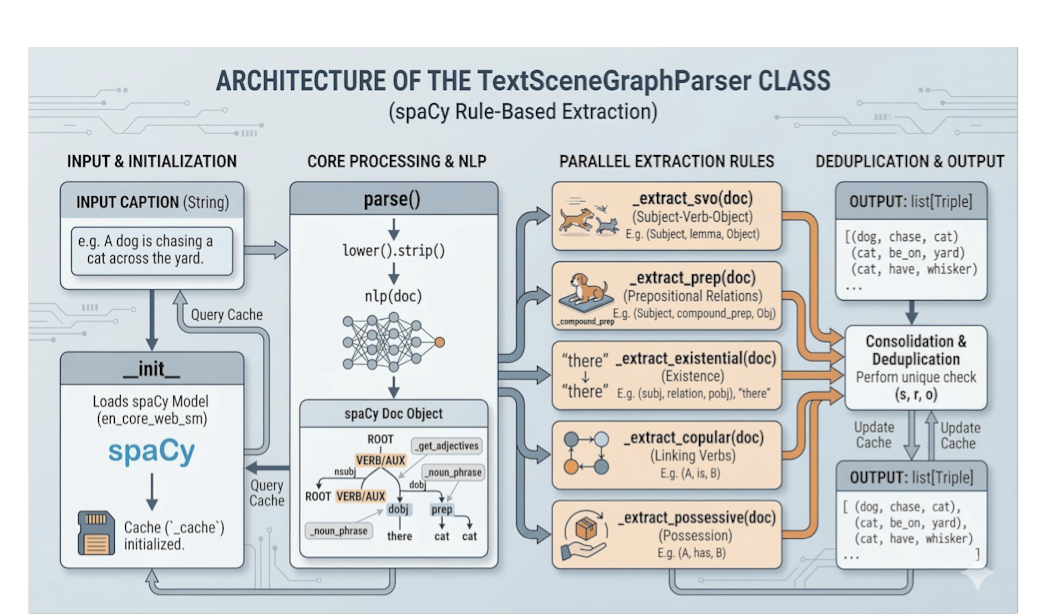}
  \caption{TextSceneGraphParser architecture (spaCy rule-based). Five parallel extraction rules produce triples that are deduplicated and cached per caption string.}
  \label{fig:parser}
\end{figure*}
\begin{figure*}[t]
  \centering
  \includegraphics[width=\linewidth]{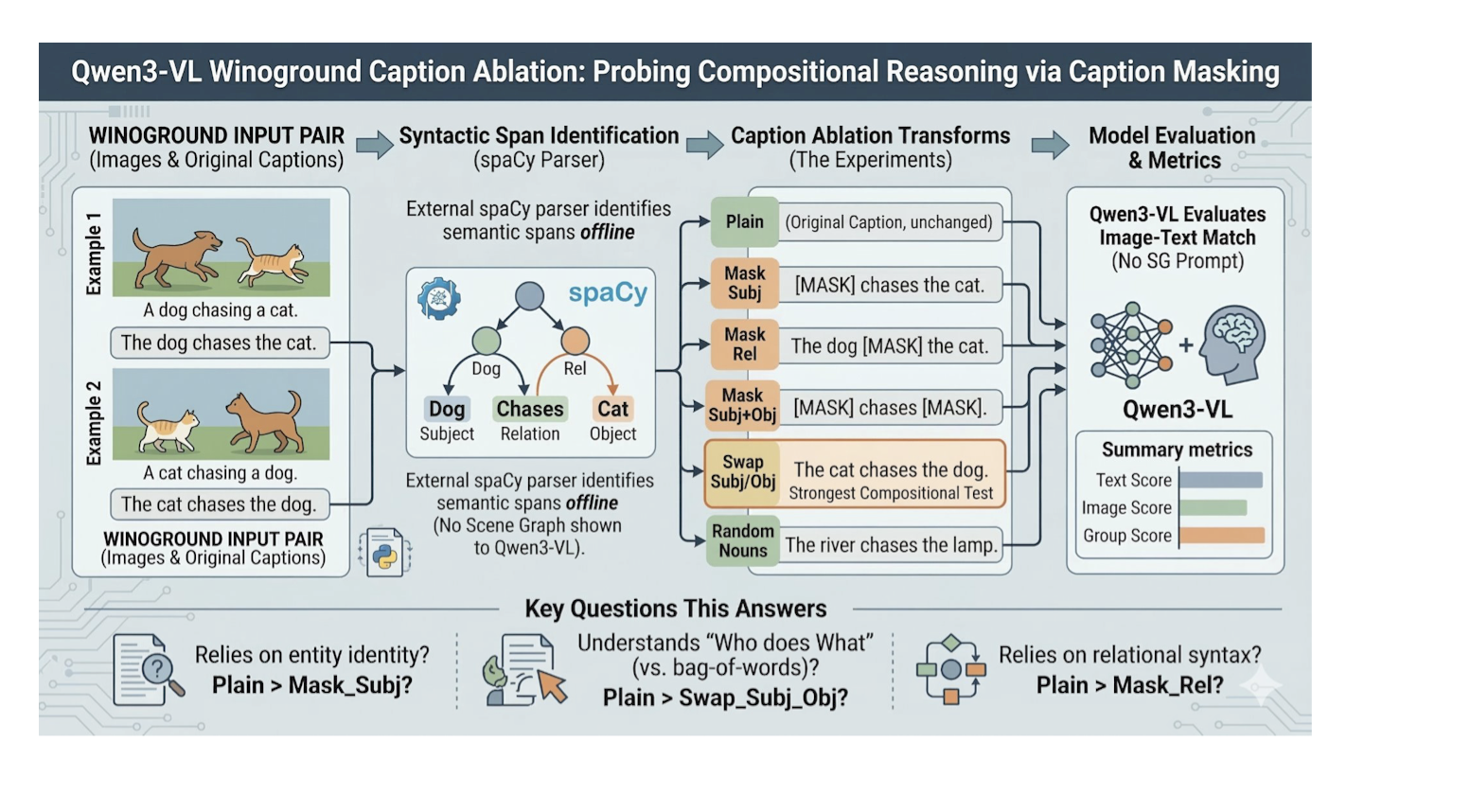}
  \caption{Caption ablation pipeline. spaCy identifies semantic spans offline; captions are transformed (mask/swap subject, object, or both) and evaluated without SG injection.}
  \label{fig:ablation}
\end{figure*}

\section{Related Work}
\label{sec:related}

\subsection{Compositional reasoning benchmarks.}
\citet{thrush2022winoground} introduced Winoground, demonstrating that all VLMs tested performed near random chance on compositional image--caption matching.
\citet{diwan2022winoground} analyzed failure modes, categorizing them as visual difficulty, linguistic ambiguity, and genuine compositional failure.
Concurrently, \citet{yuksekgonul2023and} introduced the \textbf{ARO} benchmark (Attribution, Relation, Order) with over 50,000 test cases from Visual Genome and COCO, demonstrating that contrastive VLMs systematically act as bag-of-words models , for example, BLIP assigns higher probability to ``the grass is eating the horse'' over the correct caption with 81\% confidence.
\citet{hsieh2024sugarcrepe} followed with \textbf{SugarCrepe}, identifying biases in ARO's hard negative construction and proposing LLM-generated fluent negatives across seven edit categories (replace/swap/add object, attribute, relation) that drive blind text models to chance accuracy.

\subsection{Inference-time compositional augmentation.}
A key family of methods improves compositional reasoning at inference time without retraining.
\citet{jiang2024comclip} proposed \textbf{ComCLIP}, a training-free approach that disentangles input images into subject, object, and action sub-images and performs evolving matching over compositional embeddings, mitigating spurious entity co-occurrence biases in CLIP, SLIP, and BLIP-2 across Winoground, VL-Checklist, SVO, and ComVG.
\citet{mitra2024ccot} introduced \textbf{CCoT} (Compositional Chain-of-Thought, CVPR 2024), a zero-shot prompting method that first instructs the model to generate a scene graph from the input image, then uses that graph as a structured intermediate reasoning step , improving compositional performance on Winoground, WHOOPS!, and general multimodal benchmarks without fine-tuning.
Our multi-turn SG is closely related to both: like ComCLIP we operate at inference time, and like CCoT we leverage scene graph structure, but we use \emph{text} scene graphs extracted from captions (requiring no visual SG generator) and inject them via a formally defined asymmetry scoring mechanism or multi-turn self-filtering.

\subsection{Training-time compositional fine-tuning.}
\citet{yuksekgonul2023and} proposed \textbf{NegCLIP}, fine-tuning CLIP with POS-tag-based hard negative captions, achieving substantial ARO improvements with minimal downstream degradation.
\citet{mishra2025scramble} proposed \textbf{SCRAMBLe} (ICCV Workshops 2025), generating compositional hard negatives via chain-of-thought LLM prompting and adversarial filtering, then applying direct preference optimization to Molmo-7B , achieving 54.8\% group score on Winoground, the prior open-source state-of-the-art.
\citet{herzig2023incorporating} proposed \textbf{SGVL}, incorporating Visual Genome scene graph annotations into CLIP fine-tuning via learnable graph tokens that attend to patch embeddings, improving compositional performance across multiple benchmarks.
Our method surpasses SCRAMBLe (66.0 vs.\ 54.8\%) without any retraining, positioning inference-time structural reasoning in sufficiently capable base models as a compelling alternative to preference tuning.

\subsection{Large multimodal models.}
The shift to billion-parameter generative VLMs has produced striking Winoground gains.
\citet{liu2023visual} showed LLaVA-1.5-13B achieves 36.5\% group via yes/no scoring.
\citet{lin2024evaluating} introduced \textbf{VQAScore}, using a VQA model to compute image--text alignment as the probability of ``yes'' to ``Does this image show \{caption\}?'', providing richer compositional signal than cosine similarity.
Most recently, \citet{wu2025testtime} showed that reframing evaluation as a global matching problem (GroupMatch) and applying test-time self-training enables SigLIP-B16 to reach 67\% group score and GPT-4.1 to surpass human performance (91.4\%) , revealing that standard GroupScore metrics systematically underestimate hidden model capabilities.

\subsection{Scene graphs in VL understanding.}
Scene graphs , directed graphs encoding objects, attributes, and relations , have been used for image retrieval \citep{johnson2015image}, VQA \citep{shi2019explainable}, and image generation \citep{johnson2018image}. Text scene graphs parse captions into triple structures analogously. Our TextSceneGraphParser is most closely related to spaCy-based parsers used in SGVL \citep{herzig2023incorporating} and CCoT \citep{mitra2024ccot}, but operates purely on caption text with explicit handling of five syntactic construction types.

\section{Winoground Benchmark}
\label{sec:dataset}

Each example $(I_0, I_1, c_0, c_1, \tau)$ consists of two images, two captions sharing the same content words but differing in relational structure, and a linguistic-phenomenon tag.
Given scoring function $s : \mathcal{I} \times \mathcal{C} \to \R$, we compute four scores $s_{00}, s_{10}, s_{01}, s_{11}$ and define:
\begin{align}
\text{Text}  &= \mathbb{1}[s_{00}>s_{10}]\cdot\mathbb{1}[s_{11}>s_{01}] \\
\text{Image} &= \mathbb{1}[s_{00}>s_{01}]\cdot\mathbb{1}[s_{11}>s_{10}] \\
\text{Group} &= \text{Text} \cdot \text{Image}
\end{align}
Random-chance baselines: Text = Image = 0.25, Group $\approx 0.063$.

\section{Models and Scoring Functions}
\label{sec:models}

We evaluate four VLMs spanning the major architectural paradigms (Table~\ref{tab:scoring_summary}).

\begin{table}[h]
\centering
\caption{Base scoring functions per model.}
\label{tab:scoring_summary}
\small
\begin{tabular}{@{}lll@{}}
\toprule
\textbf{Model} & \textbf{Architecture} & \textbf{Scoring} \\
\midrule
CLIP ViT-B/32  & Dual encoder          & CLS cosine \\
BLIP ITM Base  & Cross-attention       & $P(\text{match})$ \\
LLaVA 1.5-7B   & Generative decoder    & $P(\text{yes})$ \\
Qwen3-VL-8B    & Generative decoder (MoE) & EOS cosine \\
\bottomrule
\end{tabular}
\end{table}

\textbf{CLIP} \citep{radford2021learning} uses cosine similarity between $\ell_2$-normalised CLS-token embeddings: $s_\text{CLIP}(I,c) = \frac{1}{2}(1 + \cosim(f_\text{img}(I), f_\text{txt}(c)))$.

\textbf{BLIP} \citep{li2022blip} uses a cross-attention ITM head: $s_\text{BLIP}(I,c) = \softmax([v_\text{no}, v_\text{yes}])_1$.

\textbf{LLaVA} \citep{liu2023visual} scores via yes/no logits after an image-grounded prompt: $s_\text{LLaVA}(I,c) = \exp(z_\text{yes})/(\exp(z_\text{yes})+\exp(z_\text{no}))$.

\textbf{Qwen3-VL-8B-Thinking} \citep{qwen3vl} uses EOS-token pooling from a generative Mixture-of-Experts decoder, scored as CLIP but with richer contextual representations from grounding training.

\section{TextSceneGraphParser}
\label{sec:parser}

Figure~\ref{fig:parser} shows the parser pipeline. A text scene graph for caption $c$ is:
$\mathcal{G}(c) = \{\triplet{s_k}{r_k}{o_k}\}_{k=1}^K$
where $s_k, o_k$ are noun-phrase entities and $r_k$ a relation (verb, preposition, or copula). Noun phrases are enriched with adjectival and compound modifiers.

Five rules operate on spaCy's dependency tree \citep{honnibal2020spacy}: \textbf{(R1)~SVO} , transitive/intransitive verbs with relative/subordinate clause handling and negation; \textbf{(R2)~Prepositional} , spatial/temporal relations via \texttt{prep}/\texttt{pobj} arcs; \textbf{(R3)~Existential} , ``there is/are'' constructions; \textbf{(R4)~Copular} , linking-verb sentences; \textbf{(R5)~Possessive} , ``has'' triples from \texttt{poss} arcs. Invalid triples are discarded and results cached for efficiency.

\section{Graph Asymmetry Scoring}
\label{sec:gas}

Given triples $t_a = \triplet{s_a}{r_a}{o_a}$ and $t_b = \triplet{s_b}{r_b}{o_b}$ with text embedding function $\phi$, the \textbf{pairwise asymmetry} is:
\begin{equation}
A(t_a, t_b) = \alpha\bigl(\sigma(s_a,s_b) - \sigma(s_a,o_b)\bigr) + \gamma\bigl(\sigma(o_a,o_b) -
\sigma(o_a,s_b)\bigr)
\end{equation}
where $\sigma(x,y)=\cosim(\phi(x),\phi(y))$ and $\alpha=\gamma=1.0$. The relation term cancels in the forward--flipped difference, specifically targeting agent--patient role reversal.

For graphs $\mathcal{G}_a, \mathcal{G}_b$, we build cost matrix $C_{ij}=A(t_a^i, t_b^j)$ and solve optimal bipartite matching via the Hungarian algorithm \citep{kuhn1955hungarian}:
\begin{equation}
\sgprior(\mathcal{G}_a,\mathcal{G}_b) = \frac{1}{|\text{matched}|}\sum_k C_{\pi^*(k),\sigma^*(k)}
\end{equation}
When either graph is empty, $\sgprior = 0$, ensuring graceful degradation.

\subsection{Additive augmentation (CLIP, BLIP, Qwen3).}
$s_\text{SG}(I,c;\mathcal{G}_c,\mathcal{G}_{c'}) = s_\text{base}(I,c) + \lambda\cdot\sgprior(\mathcal{G}_c,\mathcal{G}_{c'})$, with $\lambda=0.3$. Since the prior depends only on the caption pair, \emph{image scores are invariant}; only text and group scores can change.

\subsection{Prompt injection (LLaVA).}
Triples are injected into the natural-language prompt, instructing the model to attend to agent--patient structure before the yes/no decision.

\section{Caption Ablation Study}
\label{sec:ablation}

To mechanistically probe what information models rely on, we apply systematic caption transforms (Figure~\ref{fig:ablation}) and measure group-score drop $\Delta_\text{grp}$ (Table~\ref{tab:ablation}).

\begin{table*}[t]
\centering
\caption{Caption ablation study across all four models. Each cell shows group score / $\Delta_\text{grp}$ vs.\ original caption (negative = degradation). Text and Image scores are also reported for completeness.}
\label{tab:ablation}
\small
\setlength{\tabcolsep}{5pt}
\begin{tabular}{@{}l ccc ccc ccc ccc@{}}
\toprule
& \multicolumn{3}{c}{\textbf{CLIP ViT-B/32}} & \multicolumn{3}{c}{\textbf{BLIP ITM}} & \multicolumn{3}{c}{\textbf{LLaVA 1.5-7B}} & \multicolumn{3}{c}{\textbf{Qwen3-VL-8B-T}} \\
\cmidrule(lr){2-4}\cmidrule(lr){5-7}\cmidrule(lr){8-10}\cmidrule(lr){11-13}
\textbf{Condition} & \textbf{Text} & \textbf{Img} & \textbf{Grp} & \textbf{Text} & \textbf{Img} & \textbf{Grp} & \textbf{Text} & \textbf{Img} & \textbf{Grp} & \textbf{Text} & \textbf{Img} & \textbf{Grp} \\
\midrule
Random Chance        & .250 & .250 & .063 & .250 & .250 & .063 & .250 & .250 & .063 & .250 & .250 & .063 \\
Original Caption     & .308 & .113 & .090 & .485 & .240 & .200 & .453 & .368 & .275 & .745 & .720 & .628 \\
\midrule
Mask Subjects        & .243 & .118 & .078{\small/$-$.013} & .370 & .213 & .158{\small/$-$.043} & .350 & .250 & .180{\small/$-$.095} & .573 & .543 & .450{\small/$-$.178} \\
Mask Objects         & .163 & .095 & .053{\small/$-$.038} & .258 & .178 & .133{\small/$-$.068} & .205 & .173 & .090{\small/$-$.185} & .420 & .460 & .338{\small/$-$.290} \\
Mask Subj+Obj        & .110 & .060 & .025{\small/$-$.065} & .135 & .120 & .073{\small/$-$.128} & .115 & .095 & .045{\small/$-$.230} & .270 & .305 & .223{\small/$-$.405} \\
Swap Subj$\leftrightarrow$Obj & .248 & .103 & .075{\small/$-$.015} & .403 & .180 & .153{\small/$-$.048} & .360 & .290 & .203{\small/$-$.073} & .535 & .525 & .430{\small/$-$.198} \\
\bottomrule
\end{tabular}
\end{table*}

\begin{table}[t]
\centering
\caption{Winoground results. Best per-model in \textbf{bold}. Random: Txt/Img = 0.25, Grp = 0.063.}
\label{tab:main}
\small
\setlength{\tabcolsep}{4pt}
\begin{tabular}{@{}llccc@{}}
\toprule
\textbf{Strategy} & \textbf{Model} & \textbf{Txt} & \textbf{Img} & \textbf{Grp} \\
\midrule
Random            & {,}          & .250 & .250 & .063 \\
\midrule
\texttt{clip}     & CLIP ViT-B/32  & .308 & .113 & .090 \\
\texttt{clip+SG}  & CLIP ViT-B/32  & \textbf{.308} & \textbf{.113} & \textbf{.090} \\
\midrule
\texttt{blip}     & BLIP ITM       & .485 & .240 & .200 \\
\texttt{blip+SG}  & BLIP ITM       & \textbf{.485} & \textbf{.240} & \textbf{.200} \\
\midrule
\texttt{llava}    & LLaVA 1.5-7B   & \textbf{.453} & \textbf{.368} & \textbf{.275} \\
\texttt{llava+SG} & LLaVA 1.5-7B   & .365 & .308 & .203 \\
\midrule
\texttt{qwen3}         & Qwen3-8B-T & .745 & .720 & .628 \\
\texttt{qwen3+SG}      & Qwen3-8B-T & .745 & .735 & .642 \\
\texttt{qwen3+SG (mt)} & Qwen3-8B-T & \textbf{.760} & \textbf{.720} & \textbf{.660} \\
\bottomrule
\end{tabular}
\end{table}

\begin{table}[t]
\centering
\caption{Comparison with published methods on Winoground (\%).}
\label{tab:sota}
\small
\setlength{\tabcolsep}{4pt}
\begin{tabular}{@{}lccc@{}}
\toprule
\textbf{Method} & \textbf{Text} & \textbf{Image} & \textbf{Group} \\
\midrule
MTurk Human              & 89.5 & 88.5 & 85.5 \\
Random Chance            & 25.0 & 25.0 &  6.3 \\
\midrule
CLIP (ViT-B/32)          & 30.8 & 10.5 &  8.0 \\
BLIP (ITM Large)         & 35.0 & 22.0 & 22.0 \\
LLaVA-1.5-13B            & 51.5 & 50.5 & 36.5 \\
CECE (LLaVA-1.5+1.6)     & 55.0 & 61.3 & 47.5 \\
Molmo-7B                 & 62.8 & 61.8 & 49.5 \\
SCRAMBLe-Molmo-7B \citep{mishra2025scramble} & 66.8 & 66.3 & 54.8 \\
\midrule
Qwen3-VL-Think (plain)   & 74.5 & 72.0 & 62.8 \\
Qwen3-VL-Think + SG      & 74.5 & 73.5 & 64.2 \\
\textbf{Qwen3+SG (mt)}   & \textbf{76.0} & \textbf{72.0} & \textbf{66.0} \\
\bottomrule
\end{tabular}
\end{table}

Three key findings: \textbf{(1)~Qwen3-VL shows the largest absolute drops} across all conditions, confirming it relies most on entity identity and relational syntax. \textbf{(2)~Masking objects consistently hurts more than masking subjects} across all models. \textbf{(3)~The swap condition} ($\Delta \approx -0.20$ for Qwen3) confirms partial but imperfect compositional sensitivity , role reversal disrupts but does not destroy performance.

\section{Multi-Turn SG Filtering}
\label{sec:multiturn}

\begin{figure*}[t]
  \centering
  \includegraphics[width=\linewidth]{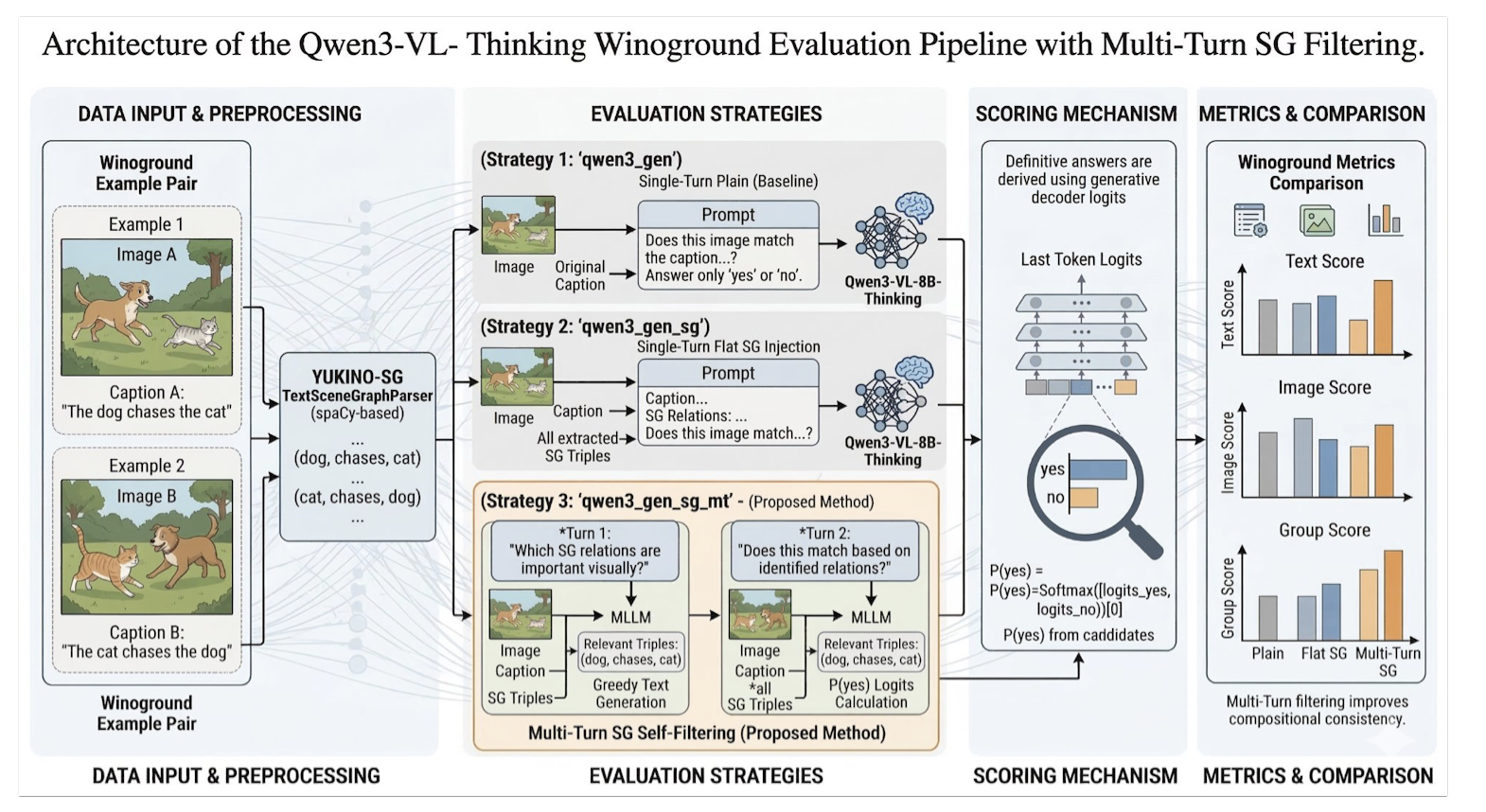}
  \caption{Qwen3-VL-Thinking pipeline with multi-turn SG filtering. Turn~1 identifies visually relevant SG relations; Turn~2 uses those filtered triples for the final match decision.}
  \label{fig:multiturn}
\end{figure*}

\begin{figure*}[h]
  \centering
  \includegraphics[width=0.7\linewidth]{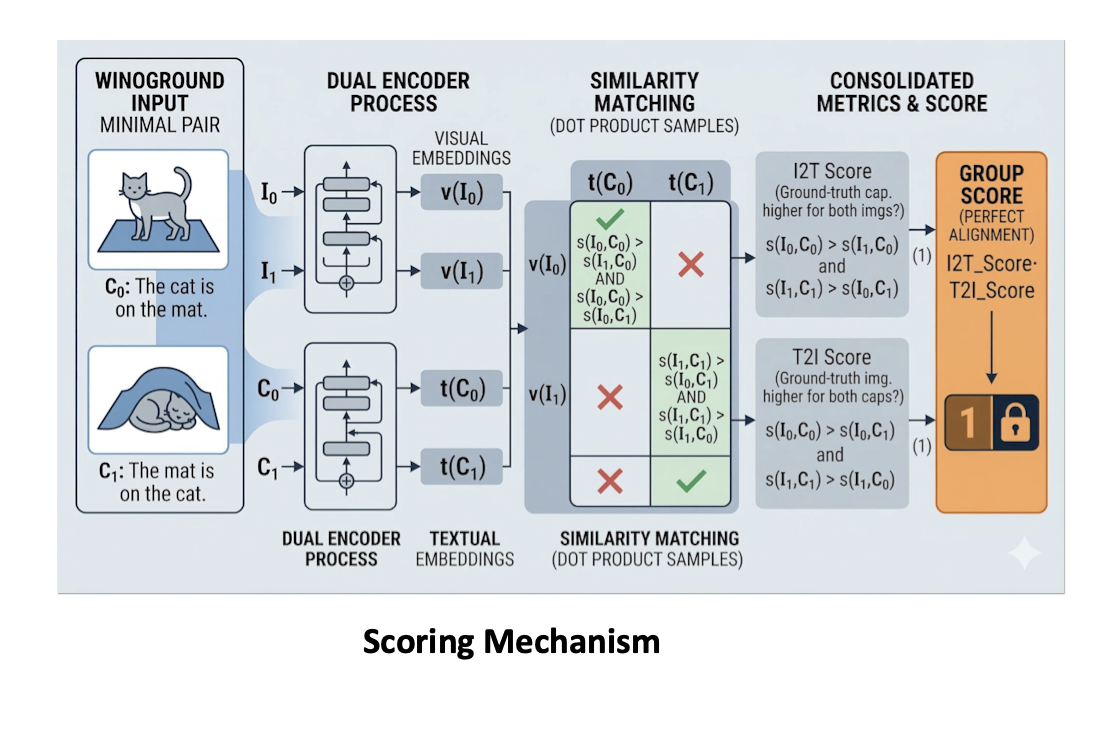}
  \caption{Scoring mechanism: last-token logits for ``yes''/``no'' are softmax-normalized to yield $P(\text{yes}\mid I, c)$.}
  \label{fig:scoring}
\end{figure*}

Figure~\ref{fig:multiturn} shows three strategies for Qwen3-VL-8B-Thinking. \textbf{(S1) Plain}: single-turn yes/no, no scene graph. \textbf{(S2) Flat SG injection}: all extracted triples in a single prompt. \textbf{(S3) Multi-turn SG filtering} (proposed): Turn~1 generates the visually relevant subset of triples via greedy decoding; Turn~2 uses only those filtered triples for the final match query (Figure~\ref{fig:scoring}). This \emph{visual grounding filter} reduces noise from parser errors and irrelevant triples.

\section{Results}
\label{sec:results}

\subsection{Main Results}

Table~\ref{tab:main} presents results for all strategies. Qwen3-VL-8B-Thinking plain (62.8\% group) already substantially outperforms all encoder-based models. Multi-turn SG filtering further improves to 66.0\%.

\subsection{Comparison with Prior Work}

Table~\ref{tab:sota} places results in context. Qwen3+SG (multi-turn) achieves 66.0\% group, surpassing SCRAMBLe-Molmo-7B (54.8\%) by 11.2 points and approaching human performance (85.5\%), without any additional training.

\section{Analysis and Discussion}
\label{sec:analysis}

\subsection{Capability--augmentation tradeoff.}
For weaker encoder-based models (CLIP, BLIP), the additive SG prior produces negligible gains , the prior's signal is too weak to overcome insufficient base compositional sensitivity. For LLaVA, prompt injection hurts ($\Delta_\text{grp}=-0.073$): the model anchors on explicitly listed entities rather than integrating visual and structural evidence jointly. By contrast, Qwen3-VL benefits from SG filtering as a \emph{compositional regulariser}, improving consistency of relational judgments. This pattern aligns with the view of \citet{mitra2024ccot} that scene graph augmentation is most effective when the base model already possesses reasonable visual grounding capability.

\subsection{LLaVA and the over-specification failure mode.}
The performance drop under SG prompt injection for LLaVA ($\Delta_\text{grp}=-0.073$) instantiates a well-documented phenomenon in structured prompting: \emph{over-specification}, wherein explicitly enumerating relational triples in the prompt causes the model to anchor on listed entities and suppress its own visual inference \citep{mitra2024ccot}.
Unlike Qwen3-VL, which uses scene graph information as a \emph{filter} over its existing visual representations, LLaVA treats injected triples as primary evidence, overriding the image signal.
This suggests that SG injection is only beneficial when the model has sufficient capacity to \emph{integrate} structural priors with visual evidence rather than substitute one for the other , a distinction with practical implications for prompt design in generative VLMs.

\subsection{Why multi-turn outperforms flat injection.}
Flat SG injection yields a marginal gain (+1.4 group points). The multi-turn strategy adds +1.8 more by using the model's own visual grounding to filter irrelevant triples before the final decision, directly addressing the known limitation of CCoT-style methods that injecting inaccurate SGs can hurt performance \citep{mitra2024ccot}.

\subsection{Multi-turn SG filtering as structured self-consistency.}
The multi-turn strategy can be understood as a form of \emph{structured self-consistency} \citep{wang2023selfconsistency}: Turn~1 acts as a visually-grounded filter that samples the model's belief about which relational triples are perceptually relevant, and Turn~2 conditions the final decision on this filtered intermediate.
This differs from standard self-consistency (majority voting over independent samples) in two ways: the intermediate is \emph{structured} (a subset of parsed triples rather than a free-form chain-of-thought), and the two turns share a single context window, enabling the model to condition coherently.
The +1.8 group-point gain over flat injection thus reflects reduced noise from irrelevant triples rather than additional model computation per se, connecting our approach to the broader literature on inference-time compute allocation \citep{wu2025testtime}.

\subsection{Per-tag decomposition of SG benefit.}
Aggregate group scores obscure a likely interaction between linguistic phenomenon tag and SG augmentation efficacy.
Scene graph augmentation is structurally targeted at agent--patient role reversal, and should therefore benefit \emph{relational} and \emph{action-based} Winoground tags disproportionately, while providing little signal for \emph{non-visual} tags (e.g., ``same/different'' constructions) or \emph{pragmatics} tags where the relational structure is not the discriminating feature.
A per-tag breakdown of $\Delta_\text{grp}$ under SG augmentation would sharpen the capability--augmentation claim: rather than a model-level threshold, the effective condition may be \emph{tag-type $\times$ model capability}, with SG helping capable models specifically on relational examples.
We leave this decomposition to future work but note it as an important diagnostic for any augmentation strategy on Winoground.

\subsection{Positioning relative to prior work.}
Relative to \textbf{CCoT} \citep{mitra2024ccot} we use text scene graphs (no visual SG generator required) with a formally defined asymmetry scorer, while CCoT generates visual SGs via LMM prompting. Relative to \textbf{SCRAMBLe} \citep{mishra2025scramble} we require no retraining and achieve +11.2 group points, suggesting inference-time structural reasoning in sufficiently capable models can outperform training-time preference tuning in smaller models. Our results also complement \citet{wu2025testtime}, who show that evaluation metric reframing reveals hidden compositional capability in models like GPT-4.1, while we demonstrate that structured inference-time augmentation can unlock complementary compositional gains in open-weight models.

\subsection{Limitations and the ablation as a compositional diagnostic.}
The caption ablation experiments (Section~\ref{sec:ablation}) yield a structured performance fingerprint , group score under subject masking, object masking, joint masking, and subject--object swap , that characterises each model's reliance on entity identity versus relational syntax independently of overall accuracy.
We propose this \emph{ablation diagnostic} as a reusable evaluation protocol: applied to any new VLM on Winoground, it produces a four-dimensional profile that predicts whether SG augmentation is likely to help (high swap sensitivity, moderate mask sensitivity) or hurt (low overall sensitivity, suggesting bag-of-words processing).
Beyond this, the TextSceneGraphParser relies on spaCy's statistical dependency parser, which can fail on complex Winoground sentences. Future work should explore: layer-wise probing to identify which transformer layers encode agent--patient distinctions; learned graph matching replacing the Hungarian algorithm; extension to ARO \citep{yuksekgonul2023and} and SugarCrepe \citep{hsieh2024sugarcrepe}; per-tag SG benefit decomposition; and integration of visual scene graphs for image-side structural priors.

\section{Conclusion}

We presented Multi-turn SG, a unified framework for evaluating and augmenting compositional reasoning in VLMs on Winoground. Our TextSceneGraphParser and Graph Asymmetry Scorer provide architecture-appropriate structural priors without retraining. Caption ablation experiments reveal that Qwen3-VL-8B-Thinking encodes substantial but imperfect compositional structure, and multi-turn SG filtering consistently improves over both flat injection and plain baselines. Our multi-turn strategy achieves 66.0\% group score , a new open-source state-of-the-art , and our capability--augmentation tradeoff analysis provides a principled framework for understanding when structural priors help. The ablation diagnostic introduced here offers a reusable protocol for predicting augmentation efficacy in future VLMs, and the structured self-consistency framing of multi-turn filtering opens a direction for principled inference-time compute allocation in compositional reasoning tasks.

\section*{Acknowledgements}
\small
This work was completed as a course project for COSC 189: Topics in Neurosymbolic AI at Dartmouth College. The author thanks Professor Soroush Vosoughi for guidance and feedback throughout the project. Figures were generated with the assistance of Google Gemini. Code development was supported in part by Claude (Anthropic).

\bibliographystyle{plainnat}

\end{document}